\documentclass{article}

\usepackage{arxiv}
\usepackage{enumitem}

\usepackage[utf8]{inputenc} 
\usepackage[T1]{fontenc}    
\usepackage{hyperref}       
\usepackage{url}            
\usepackage{booktabs}       
\usepackage{amsfonts}       
\usepackage{nicefrac}       
\usepackage{microtype}      
\usepackage{lipsum}
\usepackage{graphicx}
\usepackage{amsmath} 
\usepackage[
  backend=biber,
  style=numeric
]{biblatex}

\graphicspath{ {./images/} }

\title{Synthetic Electrogram Generation with Variational Autoencoders}

\author{
 Miriam Gutiérrez-Fernández \\
  Vicomtech, Basque Research and Technology Alliance \\
  Universidad Rey Juan Carlos \\
  mgutierrezf@vicomtech.org \\
  Spain 
  \And
 Karen López-Linares \\
  Vicomtech, Basque Research and Technology Alliance \\
  eHealth Group, Biogipuzkoa Health Research Institute \\
  Spain
  \And
 Carlos Fambuena-Santos \\
  ITACA Institute, Universitat Politècnica de València \\
  Spain
  \And
 María S. Guillem \\
  ITACA Institute, Universitat Politècnica de València \\
  Spain
  \And
 Andreu M. Climent \\
  ITACA Institute, Universitat Politècnica de València \\
  Spain
  \And
 Óscar Barquero-Pérez \\
  Universidad Rey Juan Carlos \\
  Spain
}

\begin{document}
\maketitle
\begin{abstract}
Atrial fibrillation (AF) is the most prevalent sustained cardiac arrhythmia, and its clinical assessment requires accurate characterization of atrial electrical activity. Noninvasive electrocardiographic imaging (ECGI) combined with deep learning (DL) approaches for estimating intracardiac electrograms (EGMs) from body surface potentials (BSPMs) has shown promise, but progress is hindered by the limited availability of paired BSPM–EGM datasets. To address this limitation, we investigate variational autoencoders (VAEs) for the generation of synthetic multichannel atrial EGMs. Two models are proposed: a sinus rhythm–specific VAE (VAE-S) and a class-conditioned VAE (VAE-C) trained on both sinus rhythm and AF signals. Generated EGMs are evaluated using morphological, spectral, and distributional similarity metrics. VAE-S achieves higher fidelity with respect to in-silico EGMs, while VAE-C enables rhythm-specific generation at the expense of reduced sinus reconstruction quality. As a proof of concept, the generated EGMs are used for data augmentation in a downstream noninvasive EGM reconstruction task, where moderate augmentation improves estimation performance. These results demonstrate the potential of VAE-based generative modeling to alleviate data scarcity and enhance deep learning–based ECGI pipelines.
\end{abstract}


\section{Introduction}

The development of robust and generalizable deep learning models for atrial fibrillation (AF) characterization faces fundamental challenges rooted in the inherent properties of atrial electrophysiological signals. AF signals have lower quality due to atria's smaller muscle mass, lower voltage potentials, and susceptibility to noise contamination from both physiological sources (such as ventricular far-field effects) and technical artifacts \cite{salinet2021electrocardiographic}. These characteristics create substantial obstacles for training deep learning architectures that require large volumes of high-fidelity data to learn meaningful patterns and achieve reliable generalization across diverse case populations.

Beyond signal quality considerations, the fundamental requirements for electrocardiographic imaging (ECGI) model development introduce additional methodological constraints. Specifically, supervised learning approaches for ECGI reconstruction need paired datasets consisting of simultaneous electrograms (EGMs) and body surface potentials (BSPMs). Acquiring such paired recordings presents considerable technical challenges in in-vivo experimental settings, as it requires synchronized multi-modal data acquisition systems and careful electrode placement protocols. This requirement becomes particularly problematic when considering the spatial characteristics of available cardiac electrical recordings.
Conventional intracavitary EGMs, while providing high temporal resolution, are inherently limited by their localized spatial coverage. These catheter-based measurements capture electrical activity from discrete points or small regions within the cardiac chambers, yielding spatially sparse representations of the underlying electrical substrate. In contrast, BSPMs reflect the integrated electrical activity projected from the entire cardiac surface to the body surface, representing a global electrical field distribution. This fundamental mismatch in spatial scales creates an irreconcilable problem: intracavitary EGMs cannot provide the comprehensive, spatially distributed ground truth necessary to establish meaningful correspondences with BSPMs. Without dense epicardial or endocardial mapping that captures the complete atrial electrical activation patterns, these localized measurements cannot serve as appropriate reference standards for training or validating ECGI reconstruction algorithms. This spatial incompatibility significantly constrains the availability of suitable training datasets for atrial ECGI research, limiting the development and validation of data-driven approaches in this domain \cite{hernandez2023electrocardiographic}.

Due to these limitations, most ECGI methodologies are developed and validated using simulated datasets, where the electrical activity at each point on the cardiac surface can be explicitly modeled and controlled. Although simulation can be regarded as a baseline form of data augmentation, it provides only the variability encoded in the forward model and electrophysiological parameters. As a result, simulated data alone cannot reproduce the full morphological, temporal, and spatial variability of real atrial signals. Moreover, publicly available ECGI repositories contain only limited atrial data. For example, the EDGAR database, the main open-source reference for ECGI research \cite{aras2015edgar,aras2015edgarb}, includes a wide range of simulated and experimental datasets but is predominantly composed of ventricular recordings, with atrial datasets being comparatively scarce.

Low dataset variability limits generalization in non-invasive EGM reconstruction. Baseline strategies for addressing limited dataset size and variability were discussed, including simple oversampling techniques in which minority-class samples are duplicated via sampling with replacement. In the present chapter, we explore more advanced and principled augmentation methodologies capable of enhancing both the diversity and realism of the training data beyond what naive oversampling approaches can achieve. In this work, we present a generative modeling approach using VAEs to synthesize realistic multichannel atrial EGMs, effectively addressing the limited availability of paired BSPM-EGM data required for training robust ECGI reconstruction models. The proposed approach learns a probabilistic latent representation that enables the generation of new, diverse, and physiologically plausible EGMs, thereby enriching the dataset with variability that the deterministic simulator cannot produce.

\section{Scientific Background and Study Objectives}

The generation of synthetic electrocardiographic data has gained increasing attention in recent years. Synthetic signals can enhance the interpretability of cardiac electrical phenomena, enable the creation of large and diverse datasets, and support data anonymization, thus facilitating reproducibility, benchmarking, and open-science practices~\cite{zanchi2025synthetic}.

Deep learning offers several promising approaches for synthetic ECG generation. Among them, variational autoencoders (VAEs), generative adversarial networks, and diffusion models have been most widely explored. Recent studies demonstrate the effectiveness of encoder–decoder and vector-quantized VAE (VQ-VAE) architectures in producing high-fidelity, single-heartbeat ECG waveforms from compact latent representations. These models operate either directly on raw 1-D ECG waveforms or on 2-D, image-based representations (e.g., spectrograms, wavelet maps, or lead–time matrices), and have shown strong potential for generating physiologically plausible synthetic rhythms~\cite{liu2020using, kuznetsov2020electrocardiogram}. 

VAEs are particularly well suited for generative modeling in low-data electrophysiology because they avoid the two major pitfalls that limit generative adversarial networks (GANs) and diffusion models. First, VAEs optimize a single, likelihood-based objective that yields predictable training, whereas GANs rely on adversarial optimization that is notoriously unstable, often resulting in mode collapse and poor coverage of the data distribution. Second, VAEs maintain an explicit probabilistic latent space that supports interpretable representations and robust sample generation, unlike GANs, whose latent spaces have no principled structure. Diffusion models, while powerful in vision tasks, require very large datasets, long training schedules, and remain prone to non-convergence in small-sample settings, producing synthetic signals that neither capture the full variability nor the realism of electrophysiological data~\cite{zanchi2025synthetic}. Together, these factors make VAEs the most reliable and data-efficient choice for generative modeling in electrophysiology. 

A known limitation of VAEs, however, is posterior collapse, a degeneracy in which the latent variables become uninformative because the encoder’s distribution is driven too close to the prior. This issue can be mitigated with $\beta$-annealing, which progressively increases the Kullback--Leibler (KL) weight during training to preserve informative latent representations in the early stages.

Although ECG data augmentation has been widely explored, to our knowledge, no published work has addressed the generation of EGMs, which constitute a more complex spatio-temporal signal due to their higher spatial resolution. Generating synthetic EGMs would enable the creation of richer datasets that can be propagated through the forward problem to derive corresponding BSPMs, thus supporting data augmentation for ECGI applications.

In this study, we propose a self-supervised framework for synthetic EGM generation using VAEs for the first time, motivated by the use of these models for ECG signal synthesis~\cite{Bacoyannis22, sviridov2025conditional}. We hypothesize that VAE-generated EGMs can improve generalization of non-invasive EGM estimation using deep learning models by enriching the training distribution beyond deterministic simulations. To avoid posterior collapse, we adopt a $\beta$-VAE formulation, which balances generative flexibility with regularized latent structure.

Our contributions are: (i) learning morphology-preserving latent representations using two convolutional $\beta$-VAE with perceptual loss and annealed regularization, and (ii) generating synthetic EGMs for data augmentation in downstream noninvasive deep learning-based EGM estimation tasks under sinus rhythm and AF conditions.  

The manuscript is organized as follows: Section~\ref{sec:Dataset} presents the dataset employed for the study, followed by Section~\ref{sec:model} including a description of the architecture used, the training settings, generation strategies and evaluation metrics. Section~\ref{sec:results} reports results, and Section~\ref{sec:conclusions} provides conclusions.

This section outlines the datasets, VAE architectures, training procedures, and evaluation strategies used to generate synthetic EGMs and assess their impact on noninvasive reconstruction.

\section{Dataset}\label{sec:Dataset}

Two datasets of intracardiac EGMs were used for this study. Throughout the chapter, we refer to \textit{in-silico} EGMs as deterministic computational simulations of atrial activity obtained from biophysically detailed electrophysiology models~\cite{rodrigo2017technical}. \textit{In-silico} EGMs were simulated at a sampling rate of 500~Hz from 2048 atrial sites and represent 2–4 seconds of activity.  \textit{VAE-generated} EGMs, in contrast, denote synthetic signals generated by the proposed VAE models.

\textbf{Dataset A - sinus only:}  This dataset consists of simulated EGMs from 19 sinus cases, derived from realistic atrial electrophysiology models. It is used to train the VAE model for sinus signal generation.  

\textbf{Dataset B – sinus + AF:}  This dataset extends Dataset A by incorporating an additional 33 AF cases, resulting in a combined dataset of 19 sinus rhythm and 33 AF recordings. It is used for training a class-conditioned VAE to generate both sinus and AF EGMs.

\section{Generative Variational Autoencoders and Electrogram Generation}\label{sec:model}

Two different VAE architectures were explored for EGM generation: one trained only on sinus rhythm data (VAE-S) with dataset A and one conditioned on rhythm class (VAE-C) with dataset B.

\subsection{Proposed Architectures}

\begin{figure}[t] 
    \centering
    \includegraphics[width=1\linewidth]{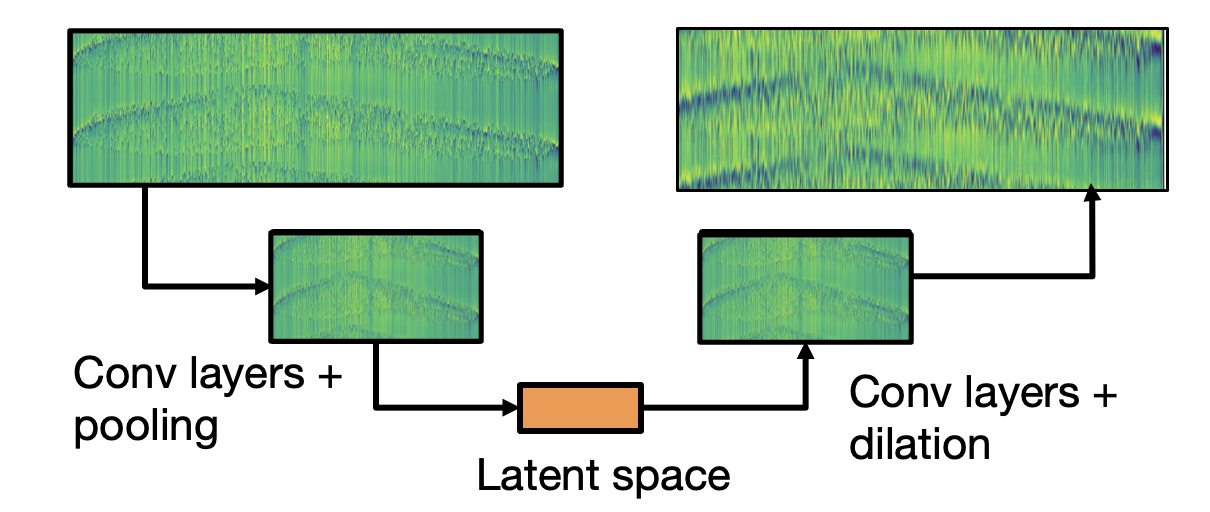}
    \caption{Architecture representation of VAE}
    \label{fig:arch}
\end{figure}

\textbf{VAE-S} was designed to model the distribution of multichannel sinus rhythm EGMs. Input signals were reshaped into 2-D tensors (\emph{time}~$\times$~\emph{channels}), with dimensions of (400, 2048), as shown in Figure~\ref{fig:arch}. The encoder consisted of successive convolutional and pooling layers that compressed the inputs into a 50-dimensional latent distribution, parameterized by mean and log-variance. Latent vectors were sampled using the reparameterization trick, a mathematical technique used in VAEs to enable gradient-based training through random nodes~\cite{kingma2013auto}.  
The decoder reconstructed EGMs from the latent space using transposed convolutions, intermediate anti-aliasing filters, and a final \emph{tanh} activation.
\textbf{VAE-C} extended the same encoder–decoder design to both sinus and AF EGMs. Conditioning was introduced by incorporating the rhythm class into the latent representation, guiding the decoder to generate class-specific reconstructions. Unlike VAE-S, VAE-C incorporates the class vector into both training and sampling, enabling class-conditioned generation.

\subsection{Training Settings}

Signals from Dataset A and Dataset B were normalized between $-1$ and $1$ and downsampled to 200~Hz to optimize batch size. Because Dataset A contains only sinus rhythm cases, it was randomly split into training (75\%), validation (15\%), and test (10\%) subsets. In contrast, Dataset B includes both sinus and AF recordings, so a stratified split was applied to preserve the class proportions in the training, validation, and test sets and ensure balanced representation of both rhythms during model training and evaluation. Both models, i.e VAE-S and VAE-C, were trained using the Adam optimizer with an initial learning rate of 0.001 and a batch size of 400 for 90 epochs, with early stopping at 10 epochs to avoid overfitting. A learning rate scheduler was applied to improve convergence. 

The VAEs were trained to minimize an objective loss function that extends beyond the standard reconstruction loss. In this work, we contributed additional perceptual terms specifically designed to preserve the morphology, local dynamics, and frequency content of in-silico EGMs. The resulting total loss objective is defined as:

\begin{equation}
\begin{split}
\mathcal{L}_{T} = & \; \mathrm{0.35} \cdot \mathcal{L}_{R} + \beta \cdot \mathcal{L}_{KL} 
+ \mathrm{0.5} \cdot \mathcal{L}_{C} + \mathrm{0.35} \cdot \mathcal{L}_{G} \\
& + \mathrm{0.25} \cdot \mathcal{L}_{H} + \mathrm{0.10} \cdot \mathcal{L}_{S}
\end{split}
\end{equation}

where $\mathcal{L}_{R}$ is the mean squared error (MSE) between original and reconstructed EGMs, $\mathcal{L}_{KL}$ is the KL divergence between the approximate posterior and the prior, where the parameter $\beta$ was linearly increased from 0 to a maximum of 4.0 over the first 10 training epochs. This gradual schedule mitigated posterior collapse in early stages and progressively enforced latent space regularization. For VAE-C, the latent prior was conditioned on rhythm class using one-hot encoding.

The perceptual terms $\mathcal{L}_{C}$, $\mathcal{L}_{G}$, $\mathcal{L}_{H}$ and $\mathcal{L}_{S}$ were designed to capture complementary aspects of signal quality. 

The \textbf{correlation loss $\mathcal{L}_{C}$} was computed as one minus the average Pearson correlation coefficient between real and reconstructed signals across nodes, emphasizing morphological similarity, as:

\begin{equation}
\mathcal{L}_{C} 
= 1 - \frac{1}{N}\sum_{n=1}^{N}
\frac{\sum_{t}(x_{t,n}-\mu_{x,n})(\hat{x}_{t,n}-\mu_{\hat{x},n})}
{\sqrt{\sum_{t}(x_{t,n}-\mu_{x,n})^{2}}
\sqrt{\sum_{t}(\hat{x}_{t,n}-\mu_{\hat{x},n})^{2}} + 10^{-8}}
\end{equation}
where $\mu_{x,n}$ and $\mu_{\hat{x},n}$ are temporal means.  

The gradient loss $\mathcal{L}_{G}$ imposed a first-order temporal gradient penalty to enforce similarity in local dynamics, particularly during rapid transitions, defined as:
\begin{equation}
    \mathcal{L}_{G}
= \frac{1}{(T-1)N}\sum_{t=1}^{T-1}\sum_{n=1}^{N}
\left| (x_{t+1,n}-x_{t,n}) - (\hat{x}_{t+1,n}-\hat{x}_{t,n}) \right|
\end{equation}

The \textbf{high-frequency loss $\mathcal{L}_{H}$} encouraged the preservation of sharp deflections, such as atrial depolarizations, which are shown as rapid and sharp peaks.  Let $L_t$ and $L_p$ denote the STFT log-magnitude spectra of the true and reconstructed signals, respectively. A high-frequency mask $m_f$ is defined as $m_f = 1$ for normalized frequencies above a cutoff 0.40 for $n_\mathrm{fft}=256$  and $0$ otherwise.  
The weighted high-frequency matching term is
\[
\mathcal{L}_{H} 
= \frac{1}{FS}\sum_{f,s}
w_{f,s}\,\big| L_{p}(f,s) - L_{t}(f,s) \big|,
\]
where $w_{f,s}$ scales with the relative high frequency amplitude of the true signal and emphasizes preservation of sharp deflections such as atrial depolarizations.

The \textbf{noise suppression loss $\mathcal{L}_{S}$} penalized spurious high-frequency components absent from the original signal. A complementary weight $w^{\mathrm{spur}}_{f,s}$ identifies high frequency regions where the ground-truth amplitude is negligible. The penalty is
\[
\mathcal{L}_{S}
= \frac{1}{FS}\sum_{f,s}
w^{\mathrm{spur}}_{f,s}\,\max\!\left(L_{p}(f,s),\,0\right),
\]
which suppresses artificial high-frequency components not present in the original signal.

All loss weights were tuned empirically to maximize performance in a range between 0 and 1.

\subsection{Electrogram Generation}

The generation process involved three steps: (i) constructing a latent distribution from the training data, 
(ii) sampling new latent vectors from this distribution, and 
(iii) decoding these latent samples into synthetic EGM signals.  The process is fully unsupervised: generated signals do not originate from any particular training EGM, but instead are produced by sampling new points in the 
learned latent space.

For each training example \(x_i\), the encoder produces an approximate posterior
\[
q_{\phi}(z \mid x_i) = \mathcal{N}(\mu_i, \sigma_i^2 \mathbf{I}),
\]
where \(\mu_i\) and \(\sigma_i\) describe how the model represents the morphology of \(x_i\) in the latent space.  
To sample new signals, we must estimate the overall latent distribution of the data, known as the aggregated posterior:
\[
q(z) = \mathbb{E}_{x \sim \text{train}}\, q_{\phi}(z \mid x).
\]

Since this distribution is not available in closed form, we approximate it  by fitting a multivariate Gaussian to the set of encoded means $\mu_i$ from the training set:
\[
z \sim \mathcal{N}(\mu_q, \Sigma_q),
\qquad
\mu_q = \frac{1}{N}\sum_{i=1}^{N} \mu_i,\qquad
\Sigma_q = \mathrm{Var}(\mu_i).
\]

Then, new VAE-generated EGMs are produced by sampling latent vectors
\[
z^{(j)} \sim \mathcal{N}(\mu_q, \Sigma_q),
\]
and decoding them as
\[
\hat{x}^{(j)} = p_{\theta}(x \mid z^{(j)}).
\]
For the VAE-C, generation additionally requires specifying a rhythm class \(c\in\{\text{sinus}, \text{AF}\}\).  
The decoder then receives both \(z\) and the class one-hot vector:
\[
\hat{x}^{(j)} = p_{\theta}(x \mid z^{(j)}, c).
\]

Importantly, since no external ground-truth quality metric exists for synthetic EGMs, and because sampling from a continuous latent space can produce unrealistic signals, the only principled criterion for assessing generation fidelity is similarity to the in-silico EGMs used for training. 

\begin{figure}[htp] 
    \centering
    \includegraphics[width=\linewidth]{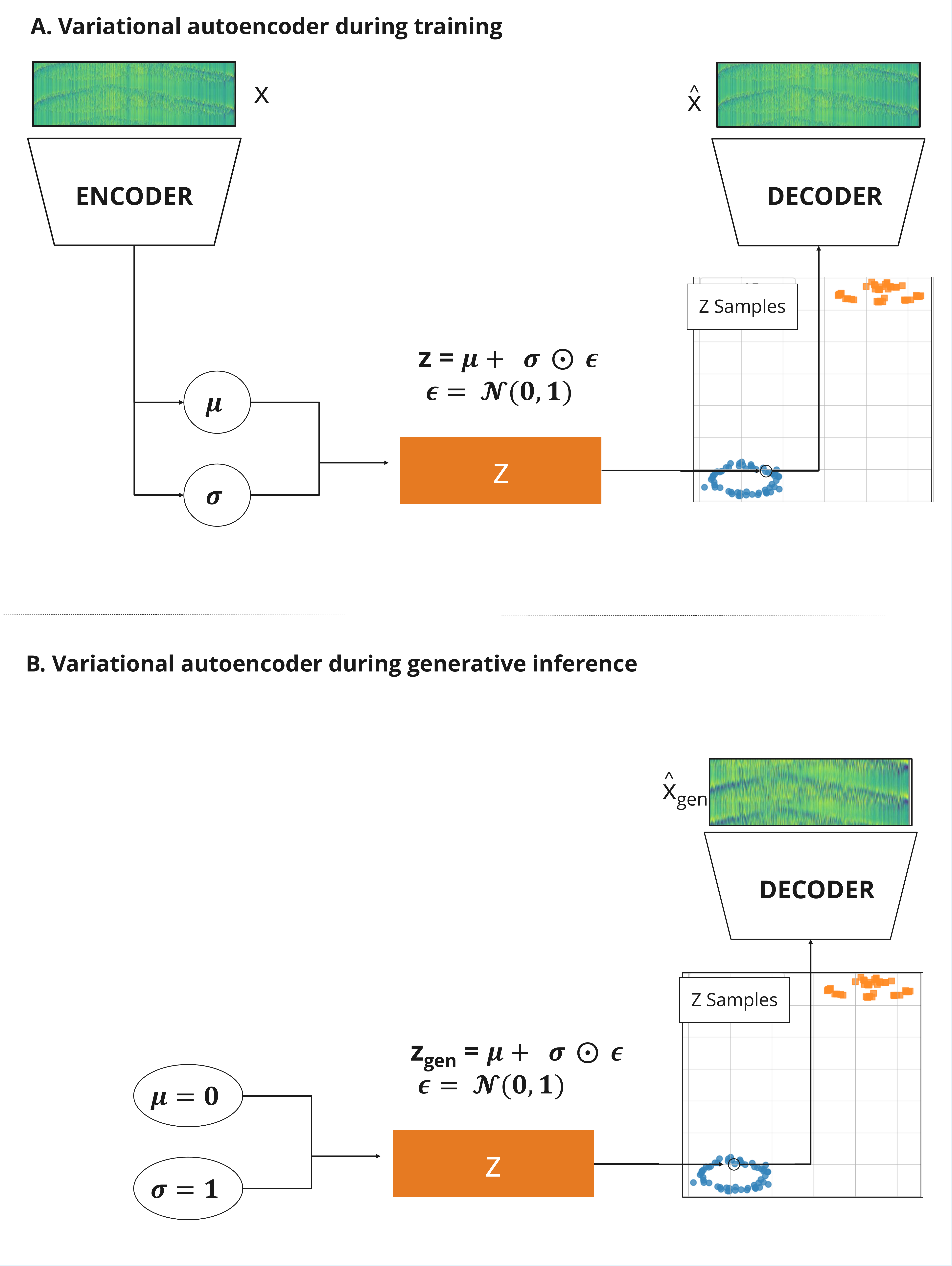}
    \caption{
    Schematic representation of a VAE.
    (A) During training, the encoder maps each input $X$ to the parameters $(\mu, \sigma)$ of a latent Gaussian distribution, from which latent samples $z$ are drawn and passed to the decoder.
    (B) During generative inference, latent variables $z$ are sampled directly from the prior $\mathcal{N}(0, I)$ and decoded to generate new data.
    }
    \label{fig:VAE_schema}
\end{figure}

Figure~\ref{fig:VAE_schema} illustrates the VAE workflow, including the encoder–decoder architecture and the latent sampling step.

For VAE-S, we generated 200 synthetic signals. Each synthetic sample was compared with all in-silico EGMs using RMSE, and the minimum RMSE across comparisons was used as its similarity score. Selecting the 25 samples with the lowest scores ensures that the final synthetic set contains only high-fidelity signals, i.e., those that most closely resemble physiologically plausible in-silico EGMs. This filtering step removes outliers or low-quality generations and retains a curated subset suitable for downstream augmentation. The 25 signals with the lowest scores were selected to form the \textbf{synt-S-Dataset}. 

For VAE-C, the same procedure was applied separately for each rhythm class. Latent vectors were sampled and decoded twice: once with the sinus label and once with the AF label. After RMSE-based filtering within each class, we retained  25 sinus-like and 25 AF-like signals, forming the \textbf{synt-C-Dataset}.  This procedure guarantees that each class subset contains realistic and morphologically consistent synthetic EGMs.

\begin{figure}[htp] 
    \centering
    \includegraphics[width=1\linewidth]{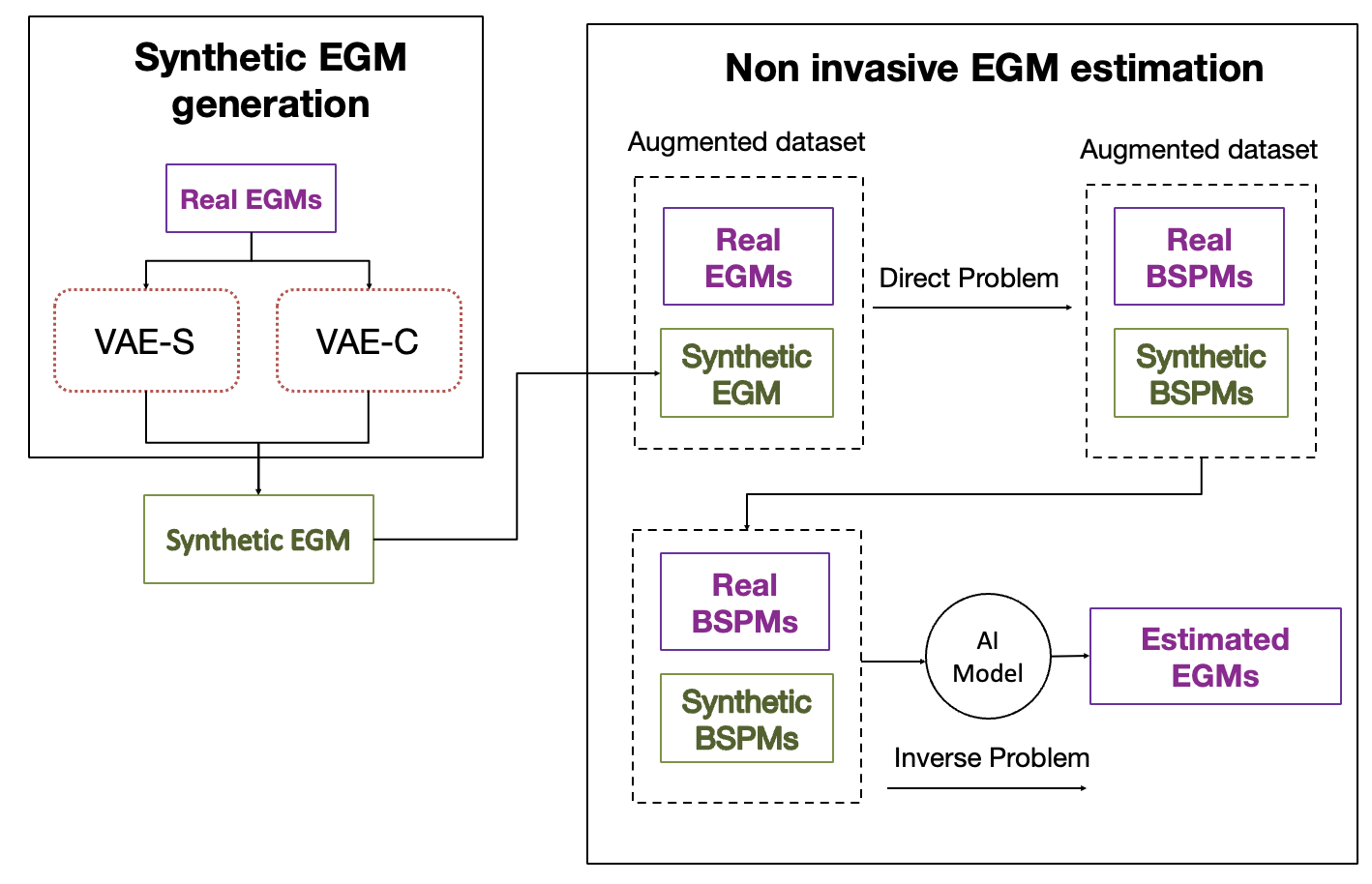}
    \caption{Overview of the complete pipeline. 
    (1) Synthetic EGMs are generated using VAE-based models. 
    (2) These generated signals are incorporated into the training data to augment the noninvasive EGM reconstruction pipeline (ECGI).}
    \label{fig:exp}
\end{figure}

The overall workflow for EGM generation and its integration into the ECGI downstream task is summarized in Figure~\ref{fig:exp}.

\subsection{Evaluation}\label{sec:eva}

We evaluated VAE performance using two complementary analyses:

\begin{enumerate}[label=(\roman*)]

   \item intrinsic fidelity metrics computed on the held-out test split, i.e., in-silico EGM signals not seen during VAE training, to quantify reconstruction quality and generative realism; 
   \item evaluation of their utility for dataset augmentation in a downstream noninvasive EGM reconstruction task.
\end{enumerate}

\textbf{Intrinsic Fidelity metrics}

To assess the similarity between VAE-generated and in-silico EGM signals, we employ metrics capturing both sample-level similarity and distributional alignment. At the reconstruction level, we report:
\begin{itemize}
    \item \textbf{MSE} which measures pointwise amplitude differences and decreases as waveform morphology becomes more similar.
    \item \textbf{KL divergence}, which quantifies mismatch between the latent distributions of real and generated samples, increasing when the VAE fails to align their statistical structure. Its value can range between 0 and $\infty$. Lower KL, indicates that the generative model aligns its latent structure with that of the training distribution.
    \item \textbf{Log-spectral distance (LSD)} which evaluates frequency-domain similarity and decreases when the spectral content (including dominant frequencies and high-frequency components) is preserved. Its value can range between 0 and $\infty$. Lower values, meaning that both low-frequency morphology and high-frequency components are faithfully reproduced~\cite{rabiner1993fundamentals}.
    \item \textbf{Pearson correlation}, which measures linear morphological similarity between signals, with values approaching 1 indicating highly correlated waveforms.
    \item  \textbf{Maximum mean discrepancy (MMD)} to assess distributional similarity where smaller values indicate that the global distribution of VAE-generated signals is closer to the distribution of in-silico EGMs. Its value can range between 0 and $\infty$. Smaller values means that the global distribution of VAE-generated EGMs closely matches the distribution of in-silico EGMs~\cite {gretton2012kernel}.
\end{itemize}

Finally, to qualitatively assess coverage and separation in the learned latent space, we project both in-silico and VAE-generated samples using t-distributed stochastic neighbor embedding (t-SNE).

\textbf{Downstream Task:} Noninvasive EGM estimation was performed using the deep learning pipeline in~\cite{gutierrez25deep}.  First, dataset B was split into training (75\%), validation (15\%), and test (10\%) sets using a stratified split (class-wise). To ensure consistency across experiments, the same test samples used to evaluate VAE generation quality were also assigned to the test split in the downstream task, so that both evaluations were conducted on an identical subset of signals. To assess the impact of VAE-generated data, we defined two augmentation scenarios for the training set. In the first ($VAE-S@k$), Dataset~B was augmented with $k$ VAE-generated sinus rhythm signals from the synt-S-Dataset, with $k \in \{10,14,18,20,25\}$. In the second (\textbf{$VAE-C@k_{S}+k_{AF}$}), Dataset~B was augmented with $k_{S}$ VAE-generated sinus and $k_{AF}$ VAE-generated AF signals from the synt-C-Dataset, where $k_{S},k_{AF} \in \{10,14,18,20,25\}$. Two settings were explored, i) targeted augmentation of $k_{S}$, only augmenting sinus rhythm, ii) symmetric augmentation of $k_{S}$ and $k_{AF}$, including both sinus rhythm and AF instances. Then, VAE-generated EGMs from the synt-S and synt-C datasets were projected to BSPMs via the forward model, producing paired BSPM–EGM samples that were concatenated with in-silico data in the training set. To prevent bias, in-silico and VAE-generated signals were alternated randomly during training. 

To evaluate the fidelity of estimated EGM signals from BSPMs, we compute Pearson correlation and RMSE between the reconstructed and ground-truth signals. 

\section{Results}\label{sec:results}

In this section, we present performance results of the EGM generation using the proposed VAE approach, in terms of generation fidelity metrics (comparing generated EGMs with in-silico ones) and impact on a downstream task, where we train an EGM reconstruction architecture with augmented dataset using the generate EGMs with VAE. 

\subsection{Electrogram Generation Performance}

This section reports fidelity metrics from EGM generation using VAE-S and VAE-C, with all results reported in Table~\ref{tab:results_gen}. For VAE-generated EGM generation with VAE-S, the model obtained a mean LSD of $2.39 \pm 0.32$ and a correlation of $0.56 \pm 0.09$ with in-silico signals, confirming reasonable frequency and morphology fidelity. Distributional similarity was assessed with MMD, yielding an average score of $0.27$. The learned latent representation exhibited an average KL divergence of $0.45 \pm 0.06$, with 24 active units out of a total latent dimensionality of 50, indicating that the model effectively utilized part of the latent capacity while avoiding posterior collapse. 

\begin{figure}[t] 
    \centering
    \includegraphics[width=1\linewidth]{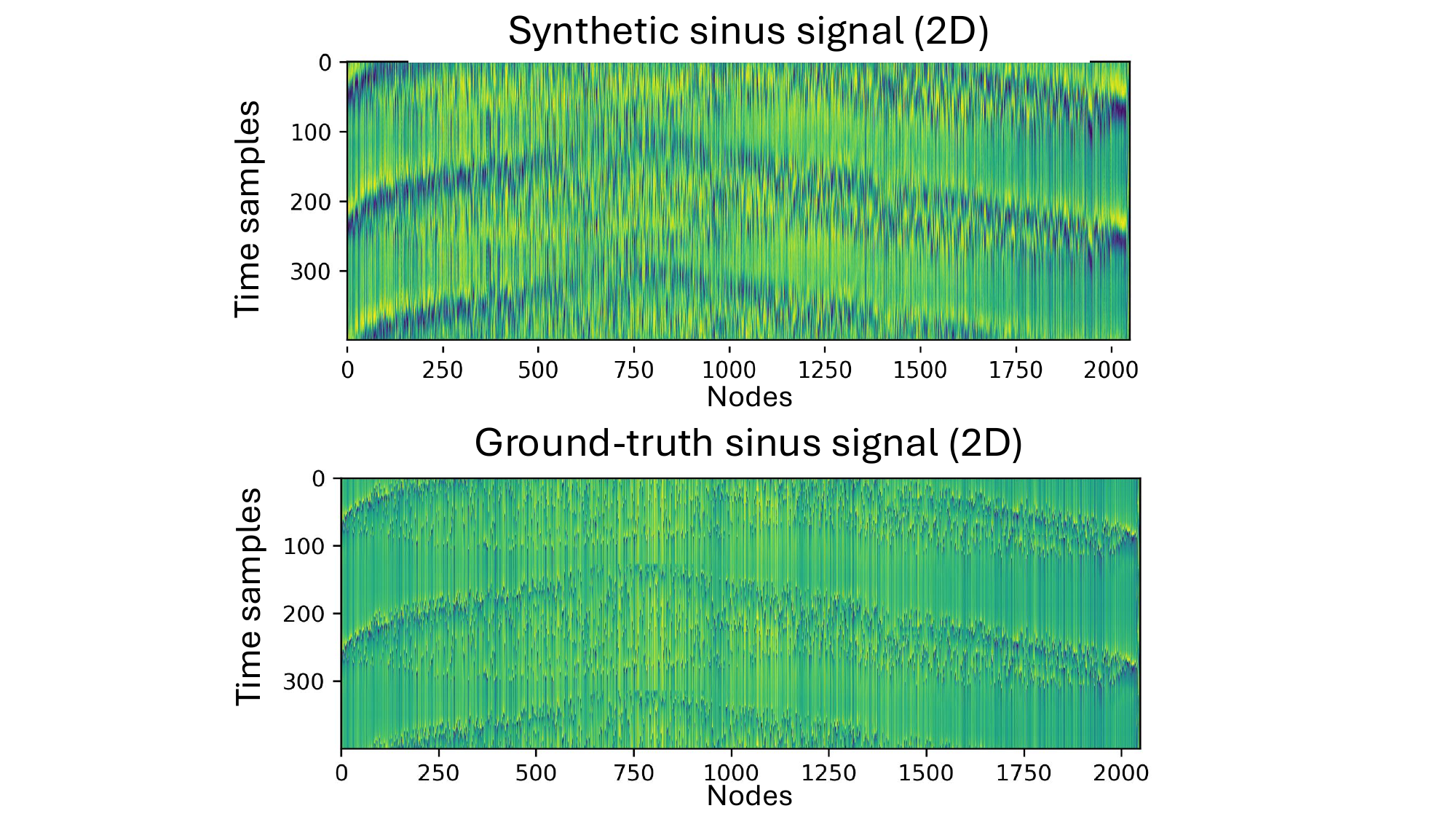}
    \caption{VAE-Generated (top) and ground-truth (bottom) signals of 400 samples, reformatted as 2-D images. X-axis: nodes (2048); Y-axis: time (400 time samples); color scale correspond to normalized amplitude: yellow=1, green=0, dark blue=-1.}

    \label{fig:VAE_sinusal_2d}
\end{figure}

\begin{figure}[t] 
    \centering
    \includegraphics[width=0.8\linewidth]{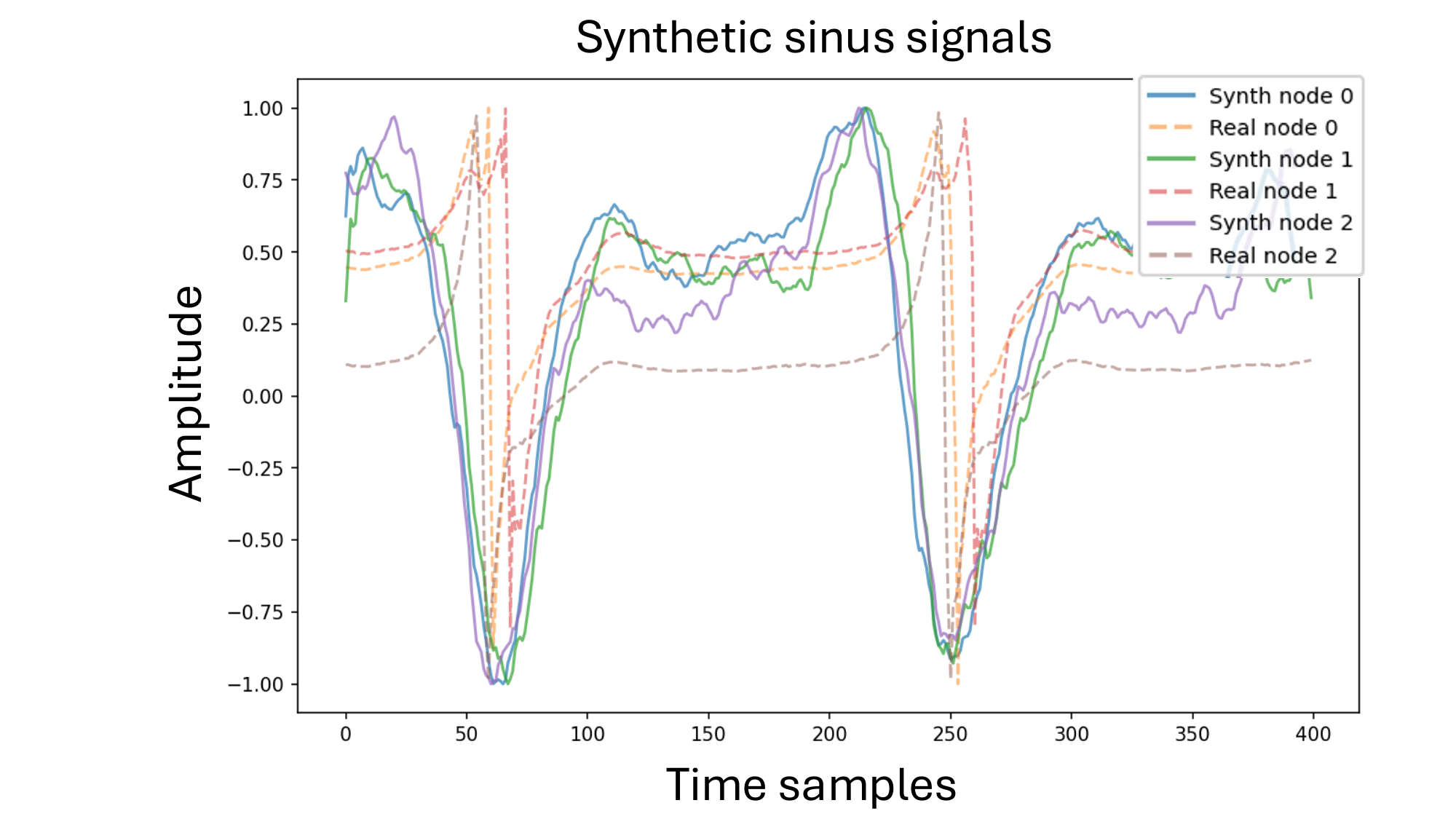}
    \caption{VAE-Generated signals produced by VAE-S. Dashed lines: in-silico signals (reference). Solid lines: VAE-generated signals. The x-axis represents time samples (corresponding to 2 seconds), the y-axis to normalized amplitude.}
    
    \label{fig:VAE_sinusal_1D_examples}
\end{figure}

Some examples of generation as 1-D signals in Fig.~\ref{fig:VAE_sinusal_1D_examples} and 2-D~\ref{fig:VAE_sinusal_2d}.

\begin{table}[ht]
\centering
\small
\caption{Fidelity metrics for VAE-S and VAE-C. Lower is better for LSD and MMD; higher is better for Correlation.}
\label{tab:results_gen}
\begin{tabular}{lcc}
\toprule
\textbf{Metric} & \textbf{VAE-S (Sinus)} & \textbf{VAE-C (Class-cond.)} \\
\midrule
LSD $\downarrow$ & $2.39 \pm 0.32$ & $2.66 \pm 0.39$ \\
Correlation $\uparrow$ & $0.56 \pm 0.09$ & $0.13 \pm 0.06$ \\
MMD $\downarrow$ & $0.27$ & Sinus: $0.56$;\; AF: $0.19$ \\
KL divergence & $0.45 \pm 0.06$ & $3.35 \pm 1.68$ \\
Active units & $24/50$ & $33/50$ \\
\bottomrule
\end{tabular}
\end{table}

\begin{figure}[t] 
    \centering
    \includegraphics[width=0.8\linewidth]{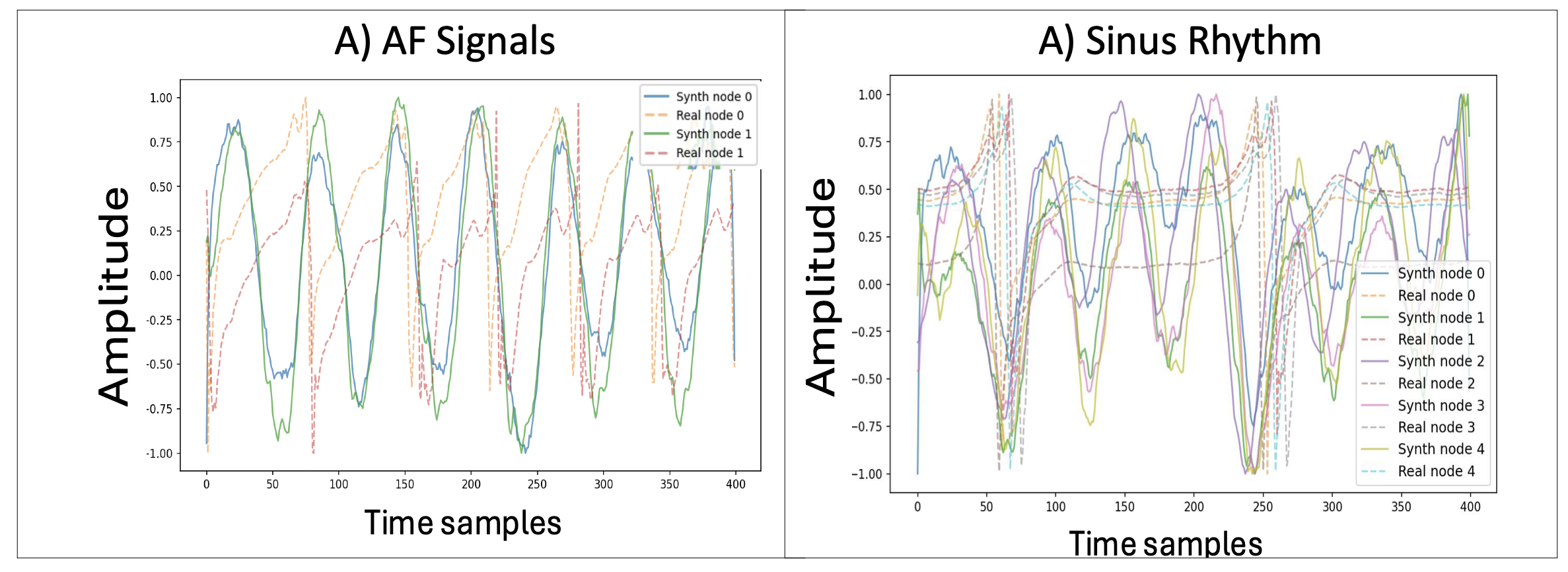}
    \caption{VAE-Generated signals produced by VAE-C of A) AF case b) Sinus rhythm case. Dashed lines: in-silico signals (reference). Solid lines: VAE-generated signals. The x-axis represents time samples (corresponding to 2 seconds), the y-axis to normalized amplitude.}
    
    \label{fig:VAE_AF_1D_examples}
\end{figure}

\begin{figure}[t] 
    \centering
    \includegraphics[width=0.35\linewidth]{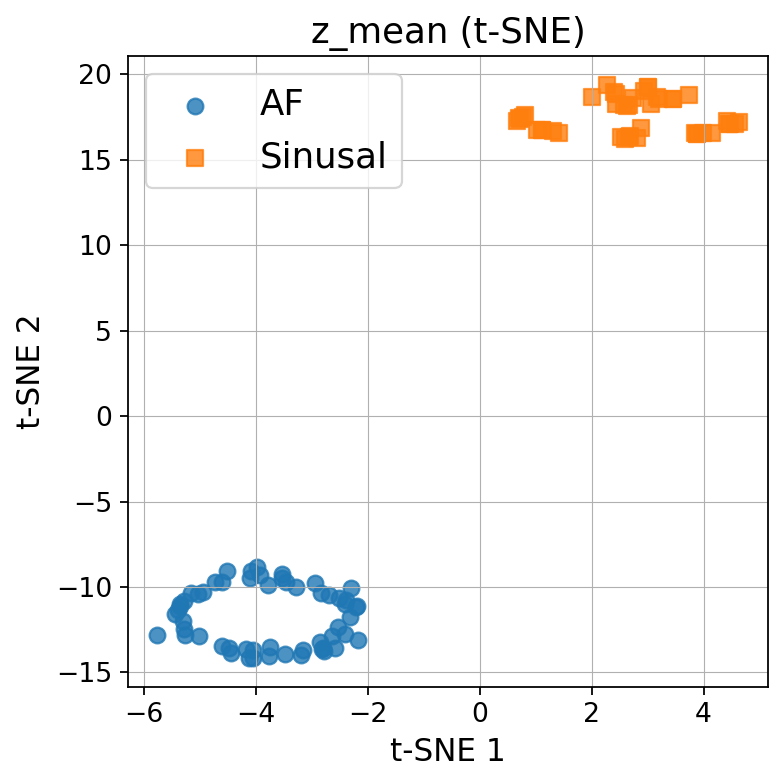}
    \caption{t-SNE projection showing two distinct clusters corresponding to AF and sinus classes }
    \label{fig:tsne}
\end{figure}

For VAE-generated EGM generation with VAE-C, the model achieved a global LSD of $2.66 \pm 0.39$ and correlation of $0.13 \pm 0.06$, indicating moderate frequency- and morphology-level fidelity. For AF, distributional similarity was stronger (MMD $0.19$) than for sinus (MMD $0.56$), suggesting more consistent generative performance in the AF domain. The t-SNE projection of the latent space, shown in Fig.~\ref{fig:tsne} revealed a clear separation between AF and sinus rhythm representations. 


\subsection{Downstream Task}

In this subsection, we present the downstream task results obtained after augmenting the training data with VAE-generated signals.

The results for $VAE-S@k$ (Table~\ref{tab:vae_all_results}) show that adding VAE-generated sinus EGMs improved reconstruction over the baseline ($k=0$). Gains were most evident with moderate augmentation ($k=14$–$18$), while larger sets ($k=20$–$25$) plateaued or slightly degraded, suggesting redundancy or bias in the training data.

\begin{figure}[ht] 
    \centering
    \includegraphics[width=0.5\linewidth]{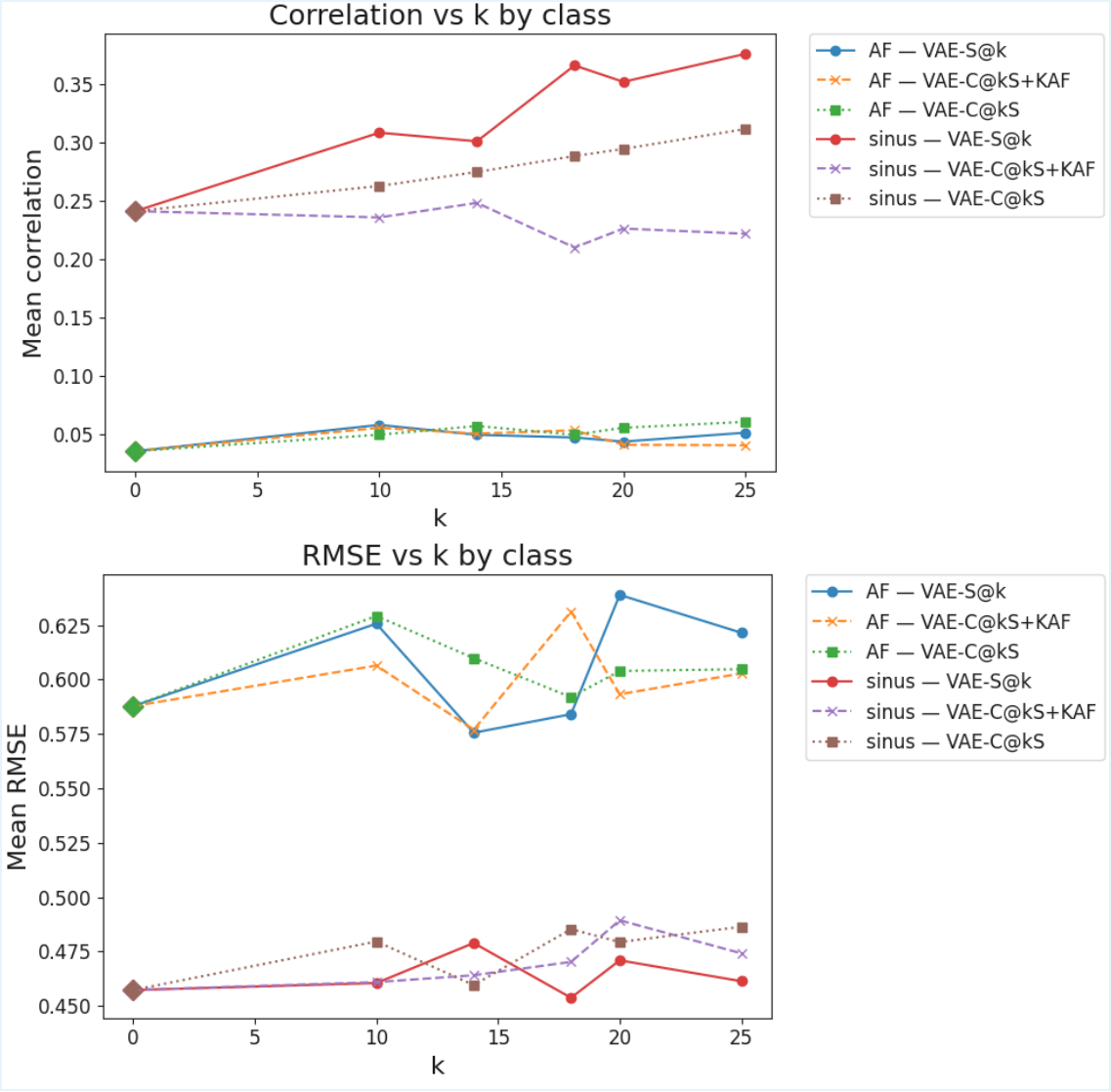}
    \caption{Comparison of mean correlation and RMSE in EGM estimation for sinus rhythm and AF under various data augmentation strategies.}
    \label{fig:corr_rmse_by_class}
\end{figure}

\begin{table}[h!]
\centering
\caption{Mean correlation (Corr) and RMSE of ground truth vs estimated EGMs from BSPMs for VAE-S@$k$, VAE-C@$k_S$, and VAE-C@$k_S,k_{AF}$ (mean $\pm$ std).}
\label{tab:vae_all_results}
\resizebox{\linewidth}{!}{%
\begin{tabular}{@{}c | cc | cc | cc@{}}
\toprule
 & \multicolumn{2}{c|}{VAE-S@$k$} & \multicolumn{2}{c|}{VAE-C@$k_S$} & \multicolumn{2}{c}{VAE-C@$k_S,k_{AF}$} \\
\cmidrule(r){2-3}\cmidrule(r){4-5}\cmidrule(l){6-7}
$k$ & Corr & RMSE & Corr & RMSE & Corr & RMSE \\
\midrule
0  & 0.160 $\pm$ 0.169 & 0.559 $\pm$ 0.103 & 0.160 $\pm$ 0.169 & 0.559 $\pm$ 0.103 & 0.160 $\pm$ 0.169 & 0.559 $\pm$ 0.103 \\
10 & 0.183 $\pm$ 0.175 & 0.543 $\pm$ 0.095 & 0.156 $\pm$ 0.163 & 0.542 $\pm$ 0.086 & 0.146 $\pm$ 0.146 & 0.543 $\pm$ 0.092 \\
14 & 0.175 $\pm$ 0.177 & 0.527 $\pm$ 0.071 & 0.166 $\pm$ 0.169 & 0.527 $\pm$ 0.082 & 0.151 $\pm$ 0.155 & 0.523 $\pm$ 0.072 \\
18 & 0.207 $\pm$ 0.201 & 0.519 $\pm$ 0.082 & 0.169 $\pm$ 0.180 & 0.523 $\pm$ 0.082 & 0.132 $\pm$ 0.125 & 0.556 $\pm$ 0.086 \\
20 & 0.198 $\pm$ 0.201 & 0.555 $\pm$ 0.096 & 0.175 $\pm$ 0.175 & 0.546 $\pm$ 0.087 & 0.138 $\pm$ 0.149 & 0.521 $\pm$ 0.066 \\
25 & 0.214 $\pm$ 0.206 & 0.541 $\pm$ 0.097 & 0.186 $\pm$ 0.182 & 0.535 $\pm$ 0.088 & 0.116 $\pm$ 0.130 & 0.540 $\pm$ 0.072 \\
\bottomrule
\end{tabular}%
}
\end{table}

The results in Table~\ref{tab:vae_all_results} show mean correlation and RMSE in the test set for $VAE-C@kS$ and $VAE-C@k_{S}+k_{AF}$. Augmenting with sinus-only generations in $VAE-C@kS$ shows a gradual improvement over the baseline, with correlation increasing from 0.16 to 0.186 at $k=25$, and RMSE remaining relatively stable. In contrast, the symmetric augmentation in $VAE-C@k_{S}+k_{AF}$ did not yield consistent benefits: performance fluctuates across $k$, and correlation values remain generally lower than for $VAE-C@kS$. 

As shown in Fig.~\ref{fig:corr_rmse_by_class}, reconstruction metrics by class (rhythm) reveal that $VAE\!-\!C@k_{S}$ improves sinus performance but remains below $VAE-S@k$, which achieves the highest correlation and lowest RMSE due to its specialization. In contrast, $VAE\!-\!C@k_{S}+k_{AF}$ introduces fibrillatory variability that slightly benefits AF, though at the expense of sinus accuracy. Overall, this indicates a trade-off: VAE-S is optimal for sinus augmentation, whereas VAE-C provides a slightly better balance across rhythms.

\section{Discussion}\label{sec:discussion}

The generation results highlight a clear trade-off between generative fidelity and rhythm generalization. VAE-S, trained exclusively on sinus rhythm data, achieved higher reconstruction fidelity and stronger morphological similarity, benefiting from its specialization. In contrast, VAE-C successfully learned rhythm-specific latent representations and enabled AF generation, as evidenced by class separation in the latent space, but at the cost of reduced sinus reconstruction quality.

These differences are reflected in the downstream task results. Augmentation with VAE-generated sinus EGMs consistently improved noninvasive EGM reconstruction, particularly when using VAE-S. This suggests that high-fidelity synthetic signals, even in limited quantities, can enrich the training distribution and improve model generalization. In contrast, symmetric augmentation with AF-generated signals did not yield consistent benefits, indicating that the additional variability introduced by fibrillatory patterns does not necessarily translate into improved global reconstruction performance.

The class-conditioned VAE introduces a compromise between rhythm coverage and signal realism. While it enables AF-specific synthesis, the reduced fidelity of sinus generations limits its effectiveness for augmentation in mixed-rhythm settings. These findings suggest that rhythm-specific generative models may be more effective than unified class-conditioned architectures when the target application prioritizes reconstruction accuracy.

\section{Conclusions and Limitations}\label{sec:conclusions}

We presented a VAE-based framework for generating VAE-generated atrial EGMs to address data scarcity. VAE-S achieved higher fidelity with sinus signals, while VAE-C enabled AF-specific generation with clear class separation but reduced sinus quality. This trade-off suggests that using separate networks for each rhythm type may enhance performance. 

Beyond intrinsic fidelity, our results demonstrate that VAE-generated EGMs can provide tangible benefits in downstream ECGI tasks. When integrated into the training pipeline for deep learning-based noninvasive EGM reconstruction, generated samples enhanced sinus reconstruction accuracy and modestly improved performance under AF when class-balanced augmentation was applied. This indicates that even moderately accurate generative models can enrich the training distribution in ways that help deep networks generalize better, particularly in low-data regimes where models are otherwise prone to overfitting the limited morphological diversity of simulated atrial activity. Importantly, these improvements validate the central hypothesis of this work: that learned generative variability complements deterministic biophysical simulations and contributes to more robust ECGI pipelines.

Nevertheless, this work also exposes important limitations. All in-silico EGMs in this study were themselves computational simulations; VAE-generated data therefore inherit simulator-induced biases. As a result, the reported gains should be interpreted as upper-bound estimates, and future work must assess generalizability to real intracardiac recordings. Furthermore, generative fidelity remains limited, especially for AF, where complex spatiotemporal patterns challenge latent-variable models such as VAEs, leading to smoother or less physiologically detailed reconstructions. Incorporating spatial context, or exploring more expressive generative architectures with larger dataset (e.g., diffusion models) could further strengthen the realism and utility of VAE-generated EGMs.

Overall, this work demonstrates the feasibility and potential of generative modelling for atrial EGM augmentation and establishes a foundation for future research that bridges computational electrophysiology, deep generative models, and noninvasive ECGI. 

\section*{Acknowledgments}
We gratefully acknowledge the ITACA Institute (Universitat Politècnica de València) and C.~Fambuena-Santos for providing simulated data used to develop this work.

\section*{Funding}

This work has been partially supported by: Fundaci{\'o}n Vicomtech, Ministerio de Ciencia ( PID2022-136887NB-I00) and Instituto de Salud Carlos III, and by Ministerio de Ciencia, Innovación y Universidades ( PID2020-119364RB-I00),  supported by European Union NextGenerationEU/PRTR (CPP2021-008562, CPP2023-01050 , CNS2022-135512).

\printbibliography[heading=bibintoc, title={Bibliography}]

\end{document}